\documentclass[conference]{IEEEtran}
\usepackage{graphicx}
\usepackage{amsmath}
\usepackage[ruled,vlined]{algorithm2e}
\usepackage{graphicx}
\usepackage{subfigure}
\ifCLASSINFOpdf
\else
\fi
\begin{document}
%
\title{Hallucination-minimized Data-to-answer Framework for Financial Decision-makers}

\author{\IEEEauthorblockN{Sohini Roychowdhury\IEEEauthorrefmark{1},
Andres Alvarez\IEEEauthorrefmark{1},
Brian Moore\IEEEauthorrefmark{1}, 
Marko Krema\IEEEauthorrefmark{1},
Maria Paz Gelpi\IEEEauthorrefmark{2},\\
Federico Martín Rodríguez \IEEEauthorrefmark{2},
Ángel Rodríguez \IEEEauthorrefmark{3},
José Ramón Cabrejas \IEEEauthorrefmark{3},
Pablo Martínez Serrano \IEEEauthorrefmark{3}\\
Punit Agrawal, \IEEEauthorrefmark{4}
Arijit Mukherjee \IEEEauthorrefmark{4}}
\IEEEauthorblockA{\IEEEauthorrefmark{1} Corporate Data and Analytics Office (CDAO), Accenture LLP, USA}
\IEEEauthorblockA{\IEEEauthorrefmark{2}CDAO, Accenture, Argentina}
\IEEEauthorblockA{\IEEEauthorrefmark{3}CDAO, Accenture, Spain}
\IEEEauthorblockA{\IEEEauthorrefmark{4}CDAO, Accenture, India,\\
Email: sohini.roychowdhury@accenture.com}}

\maketitle

\begin{abstract}
Large Language Models (LLMs) have been applied to build several automation  and personalized question-answering prototypes so far. However, scaling such prototypes to robust products with minimized \textit{hallucinations} or \textit{fake responses} still remains an open challenge, especially in niche data-table heavy domains such as financial decision making. In this work, we present a novel Langchain-based framework that transforms data tables into hierarchical textual “data chunks” to enable a wide variety of actionable question answering. First, the user-queries are classified by intention followed by automated retrieval of the most relevant data chunks to generate customized LLM prompts per query. Next, the custom prompts and their responses undergo multi-metric scoring to assess for hallucinations and response confidence. The proposed system is optimized with user-query intention classification, advanced prompting, data scaling capabilities and it achieves over 90\% confidence scores for a variety of user-queries responses ranging from \textit{\{What, Where, Why, How, predict, trend, anomalies, exceptions\}} that are crucial for financial decision making applications. The proposed data to answers framework can be extended to other analytical domains such as sales and payroll to ensure optimal hallucination control guardrails.
\end{abstract}


%
\IEEEpeerreviewmaketitle

\section{Introduction}
The Generative AI (or Gen-AI) domain powered by Large Language Models (LLMs) is one of the major digital industry disruptors of the past decade following the era of digital social networking and web-based content streaming \cite{1}. Although the name ``Generative AI" comes from the family of LLMs called ``Generative Pre-trained Transformer" (GPT) models that were developed for natural language generation (NLG) tasks \cite{2}, the current state of Gen-AI has evolved for image generation \cite{emu}, video generation \cite{runway}, and multi-modal data question-answering systems \cite{multi}. While LLMs have significantly reduced the entry barrier to AI due to the ease of use and simple language-based instructions \cite{entry}, the path from prototyping solutions powered by LLMs to building stable products has several roadblocks with regards to reliability, repeatability and data availability \cite{helm}. The major challenge identified around the trustworthiness of LLM responses is called \textit{hallucinations}, wherein, the LLM presents false facts and data as the real answer \cite{halu}. This phenomena has been found particularly difficult to control and specifically detrimental in analytical domains that involve decision making tasks based on the LLM responses \cite{dec}. In this work, we present a novel framework, infrastructure and modules that have been developed to enhance the reliability of LLM responses to assist the decision-makers of the financial domain. 

\begin{figure*}[ht]
  \centering
  \includegraphics[width=\linewidth, height=3in]{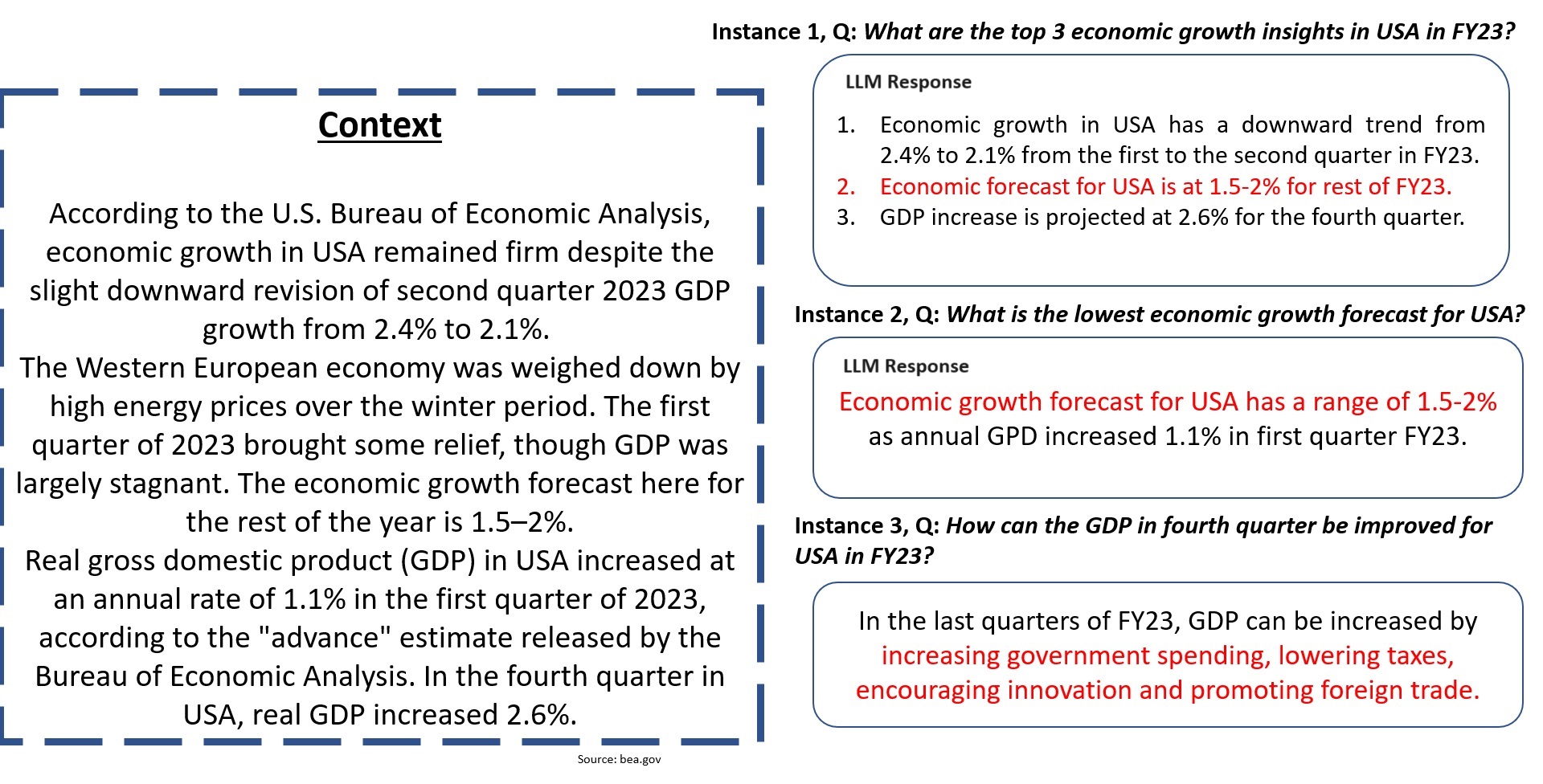}
  \caption{Examples of hallucinations by the LLMs. In this example, text chunks regarding USA and European economic growth is collected as context (left) and passed to the LLMs to answer specific questions as a custom prompt. Three user questions are tested (right) that show hallucinations by the GPT-3 (davinci-003 legacy version) LLM. All hallucinations are shown in red. For the first question, there were only 2 insights in the context, but the LLM made up three insights falsely. For the second question, the LLM focused on the smallest percentage change and hallucinated the first part of the response. For the third question, the LLM chooses to answer from past training data it has seen, which is a hallucination. }\label{hallu}
\end{figure*}

\textit{Hallucinations} predominantly occur due to the biases in the LLM training data. For instance, the early versions of the GPT-3 model that was trained on 48 TB of text data \cite{few-shot}, did not comprehend numerical or mathematical analytics and made frequent errors in simple tasks such as \textit{finding the maximum number}. Since then, the current set of LLMs available over major cloud providers have been fine-tuned and modified to better understand numerical manipulators. Some examples of hallucinations from the GPT-3 (davinci-003 legacy version) LLM available over Azure cloud are shown in Fig. \ref{hallu}. Here, we observe that even though the customizable context provided to the LLM has various named entities (such as USA, Europe etc.), LLMs can falsely switch named entities, prioritize facts with significantly low or high percentage values, or present an answer based on its prior training data as shown in the three question answering instances. This work aims to minimize such hallucinations at instruction level by building a modular framework that controls and scores for hallucinations at multiple levels, thereby allowing the users, or financial decision makers, to build trust around LLM responses.

Since most LLMs are trained on text-data, building contextual querying systems for large tabular data poses three major challenges. First, the data from tables need to be translated to language (as \textit{data chunks}) to then be queried and understood by the LLMs. Second, mathematical manipulations need to be performed in back-end analytical modules to generate data and text around trends and predictions that might be required for financial decision making purposes. It is noteworthy that LLMs have an inbuilt security layer that prohibits making predictions by default \cite{secure}, which necessitates for any predictive trends or anomalies to be separately provided to the LLM prompts for domain specific question-answering. Third, the sensibility and reliability of responses involving mathematical manipulations need to be assessed for accuracy since most LLMs are not predominantly trained on mathematical and numerical analytics. In this work, we present a novel Langchain-based \cite{langchain} framework with custom modules that control for these challenges towards a hallucination-minimized end-to-end solution.  

\begin{figure*}[ht]
  \centering
  \includegraphics[width=\linewidth, height=3.2in]{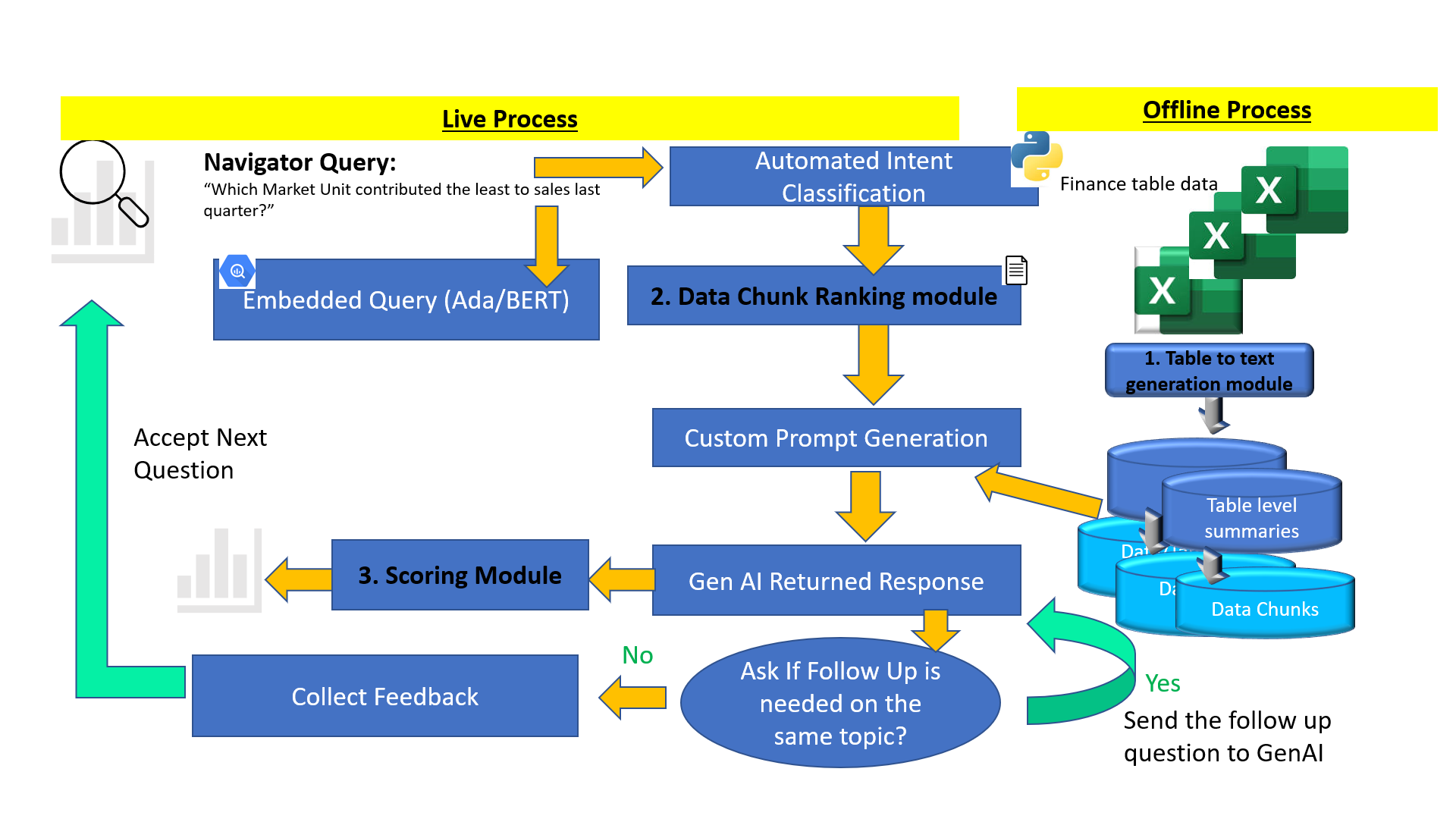}
  \caption{The proposed system architecture for the Data-to-answers framework. The offline process is triggered by data refresh rates while the online process is triggered by each user-query.}\label{sys}
\end{figure*}

This paper makes three major contributions in the form of the following \textit{hallucination-control} modules.
\begin{enumerate}
\item Table to text generation module to pass textual data to LLMs. Large data tables are converted to sentences and stored as \textit{data chunks} that are hierarchically categorized to support aggregated querying. Data chunks represent primary, secondary and trend components. Primary data chunks store the data table values as template-text while the secondary data chunks contain sentences with feature level information, i.e. which feature is the maximum etc. The trend data chunks have sentences containing predictions, anomalies and correlation information for each metric in the data tables.
\item Data chunk ranking and custom prompt generation. These components filter the most relevant data content per user-query based on the retrieval augmentation generation (RAG) \cite{RAG} mechanism and generate a customized prompt per user-query to be sent to the LLM. Filtering for data chunks followed by embedded similarity search provides a scaled framework where several hundred thousand data chunks can be mined and the most relevant data can be retrieved per user-query.
\item Live Quality Scoring module. This novel component analyzes each custom prompt the returned LLM response and evaluates the response for question context, numeric hallucinations, uniqueness from prompt, and response sensibility in terms of 6 binary quality scores. These quality scoring metrics further categorize each LLM response into \{Low/Medium/High\} confidence. The proposed framework achieves consistent response confidence scores of about 90\% and higher with iterative prompting advancements and versioned deployments. 
\end{enumerate}
The overall system diagram is shown in Fig. \ref{sys}. The proposed framework comprises of two separate processes, namely the offline and online processes. The offline process involves converting the table to text generation modules, while the live process comprises of the custom prompt generation followed by LLM response scoring per user-query.
\section{Prior work on Language Models}
Several language-models have been developed and analyzed over the past decades to enable natural language understanding (NLU) and natural language generation (NLG) tasks \cite{langhist}. All existing works can be broadly classified as encoder-only models that enable machine level understanding of human language, encoder-decoder models that enable NLU and NLG combined to process human language and generate desired outcomes such as summaries or paraphrased verbiage etc. and decoder-only models such as the LLMs. Some of the language models landmarks in each category are shown in Table \ref{NLPs}.

\begin{table}[ht] 
\caption{Landmark language models and their capabilities.}
\scalebox{0.57}
    {
\begin{tabular}{|c|c|l|}
\hline
Model Type&Example capabilities&Model names (parameters, year)\\ \hline
Encoder-only&Missing word detection&BERT \cite{bert} (110-340M, 2018), RoBERTA \cite{roberta} (125M-355M, 2019)\\
&Typing error correction&ALBERT \cite{albert} (11M-223M, 2020), ERNIE \cite{ernie} (114M, 2020)\\\hline
Encoder-decoder&Summary generation, paraphrasing&BART \cite{bart} (140M, 2019), T5 \cite{t5} (80M-11B, 2019)\\
&Limited answering&GLM \cite{glm} (110M-10B, 2021), Flan-UL2 \cite{flan} (20B, 2023)\\ \hline
Decoder-only&Language generation&GPT-1 \cite{gpt1} (110M, 2018), XLNet\cite{xlnet} (110M-340M, 2019)\\
&Conversational answering&GLam \cite{glam}(1.2T, 2021), LaMDA \cite{lambda} (137B, 2022),
Bard (137B, 2023)\\ \hline
\end{tabular}\label{NLPs}
}\vspace{-0.1cm}
\end{table}

The current generation of decoder-only LLMs with billions to trillions of parameters that are capable of maintaining conversations while presenting contextual facts such as the chatGPT \cite{gpt4} had some distinct developmental landmarks \cite{survey}. The GPT-1 model introduced in June 2018 \cite{gpt1} was the first generation of the Generative Pre-trained Transformer models and it consisted of 117 million parameters, thereby setting up the foundational architecture for the present-day ChatGPT\cite{gpt4}. The GPT-1 model demonstrated language understanding and text generation capabilities by using books as training data to predict the next word in a sentence. The GPT-2 model \cite{gpt2} that was developed in February 2019, represented a significant upgrade with 1.5 billion parameters and significant improvements in text generation capabilities by producing coherent and multi-paragraph text. However, concerns regarding unstable text generation and potential misuse prevented GPT-2 from being publicly released in early 2019. The GPT-2 model was eventually released in November 2019 after OpenAI conducted a staged rollout to study and mitigate the potential risks. The next versioned deployment of GPT-3 \cite{gpt3} was a huge improvement with 176 billion paremeters, that was launched in November 2022 by OpenAI with safety guardrails against answering questions with malefic intentions \cite{gpt3}. The GPT-3 model had far superior text-generation capabilities when compared to all other existing models at the time which led to its widespread usage in a variety of automation and question-answering applications such as drafting emails, writing articles to creating poetry and even correcting human-written programming code \cite{survey}. GPT-3 also demonstrated an ability to answer factual questions and translate between languages. Additionally, the OpenAI platform provided users the opportunity to interact with GPT-3 model through a chat-like environment (chatGPT) for free to enhance the question-answering capabilities of the model. Enhancements to the chatGPT platform with GPT3.5 and GPT-4 \cite{gpt4} further enabled automation prototypes and solutions to be developed with the LLMs. Some of the well-known foundational LLMs that are currently being applied for building automation and question-answering products are presented in Table \ref{tab:params}.

\begin{table*}[ht] 
\caption{List of Major LLMs suitable for Data to Answers use-cases and their capabilities}
\scalebox{0.88}
    {
\begin{tabular}{|c|c|l|}
\hline
 Model Name&Year& Core Capabilities\\ \hline
 GPT3.5 \cite{gpt3}&Jan 2023&- GPT 3.5 is verbose and suitable for conversational chatbots with follow up questions. \\
&&- It has variants with \{1.3, 6 and 175\} billion parameters but limited language translation and code interpretation capabilities.\\
&&- GPT-3.5 is the faster and lighter version of GPT-3 with fewer parameters (turbo) and faster response times.\\ \hline
GPT4 \cite{gpt4}&March 2023&- GPT-4 is one of the largest LLMs capable of code generation/correction, image, video and audio processing and translations.\\
&&- GPT-4 is fine tuned for multiple language translation and problem solving tasks and is the most expensive LLM per token level.\\
&&- GPT-4 is estimated to have 1.7 trillion parameters. It is fast, conversational, has least contextual limitations. \\ \hline
 Llama \cite{llama}&Feb 2023&- This open-source LLM was designed by Meta AI with 7 billion to 65 billion parameters.\\
 &&- Its major advantage is efficient language translation, summarizing and processing shorter contexts.\\
&&- LLaMA is the foundation for a variety of open-source AI models, including Dolly, Alpaca and Gorilla.\\\hline
Palm (Text/chat Bison) \cite{palm}&May 2023&-This LLM by Google contains 540 billion parameters and is stable and reliable for mathematical analytics.\\
&&-It has text and chat versions that enable conversational modalities with significantly high volumes of text processing capabilities.\\ \hline
Llama 2 \cite{llama2}&July 2023&- Llama 2 by Meta is an enhancement over LLama with enhanced safety settings against generating harmful content. \\
&&- It has 7 billion to 70 billion parameters with the ability to generate non-toxic text without additional prompting. \\ \hline
Anthropic (Calude) \cite{claude}&March 2023&- This LLM by Anthropic is trained as a conversational assistant with high ethical standards.\\
&& - Claude has 175 billion parameters and supports light browser-compatible versions along with AWS-supported virtual LLM. \\ \hline
Claude 2 \cite{claude2}&July 2023& - This LLM can generate any type of written text, summarize existing texts, and perform question-answering. \\
&& - This 130 billion parameter model has large input capacity, that can summarize hundreds of pages of documents in minutes.\\ \hline
Cohere \cite{cohere}&June 2023&- This LLM by Cohere that offers access through api calls for tasks such as summarizing, classification, and finding similarities.\\
&& - It is developed majorly for text data processing and contains 52 billion parameters. \\ \hline
BloombergGPT \cite{bloomberg}&March 2023&- This LLM contains 50 billion parameters and is trained on both domain-specific and general-purpose datasets.\\
&&- It outperforms existing LLMs on finance-specific tasks without downgraded performance on general LLM benchmarks.\\ \hline
\end{tabular}\label{tab:params}
}\vspace{-0.1cm}
\end{table*}
		
Apart from the LLMs in Table \ref{tab:params}, some other notable LLMs include Galactica that was launched by Meta in 2022 and trained on 48 million academic journals and articles \cite{galactica}. Its primary flaw was hallucinations that were difficult to detect since the language in the research articles it trained from were authoritative in nature. While most LLMs are extremely large to offer inferencing on a single laptop or system, there are some lighter versions of LLMs that run on single systems such as Orca by Microsoft \cite{orca} that contains 13B parameters and is built on top of Llama. It is noteworthy that the lighter LLM versions (with few million parameters) are characterized with limited memory and reasoning capabilities. In this work, we build an LLM-agnostic framework wherein any of the models from Table \ref{tab:params} can be applied to generate reliable and trustworthy responses on large tabular data sets.

\section{Materials and Methods}
Analytical domains such as finance, involve large volumes of tabular data that need to be accessed and returned to the user while following specific business rules. Thus, the source of our Finance-chatbot is data tables that can be pre-loaded through BigQuery. As first step in the offline process, several language data chunks are generated per-tabular record entry while maintaining the pre-defined data hierarchies such as geographical segregation at state and country levels. These data chunks represent three major types of information. The first primary-data chunks represent simple readouts of the tabular data. The second feature-level data chunks represent feature level manipulations such as minimum and maximum values per metric. The third trend-data chunks represent anomalies, exceptions and predictions returned by analytical modules that can be run offline on the data. These data chunks eliminate the burden of mathematical manipulations and the possibilities of hallucinations from LLMs. It is noteworthy that each data chunk contains 2-10 sentences of text based on the business rules. This optimal size of data chunks ensures adequate context generation for a variety of use queries. All the data chunks are subjected to vectorized embedding using Ada002 \cite{ada} or BERT embeddings \cite{bert}. The novel framework in this work aims to minimize the solution scaling challenges that appear when the number of data chunks grow over a million records in size.

The next key component for the Finance-chatbot offline process in Fig. \ref{sys} is prompt templates that contain instructions corresponding to each user question category (or intent). The categories of questions include: “What”, “Why”, “How”, “Which”, “Compare”, “Summarize”, “What-if”, “trend/anomaly/outlier” etc. In the live process of the chatbot, for every user-query, an intent classifier classifies the embedded version of the user-query to one of the existing intention categories (one vs. all classification). Based on this classified \textit{intention category}, the respective prompt template is populated with \textit{relevant data chunks} followed by scoring the response returned by the LLM for factual and contextual accuracy. The notations and descriptions of modules in the offline and live processes to generate a customized prompt per user-query are described in the following subsections.

\subsection{Notation}
The $i$-th user-query ($Q_i$) is a combination of question intention category ($I_i$) and named entities ($Q_{N_i}$) such that the named entities can be categorized as financial metric ($Q_{M_k}$), geo-location ($Q_{G_k'}$) and time-period ($Q_{T_k''}$) as shown in (1-2). The complete set of text data chunks ($C$) contain primary, feature-level and trend data chunks such that the \textit{most relevant} data chunks filtered and matched to each user-query are represented by $c_i$ and $c_{opt,i}$, respectively. The customized prompt ($P_i$) that is generated per user-query contains prompt template corresponding to intention category ($\Psi_i$), related custom definitions ($d_i$), and optimal data chunks ($c_{opt,i}$) in (3). Finally, each LLM response ($R_i$) is evaluated for the financial metric ($R_{M_k}$), geo-location ($R_{G_k'}$) and time-period ($R_{T_k''}$) in its content to return qualitative scores $S=\{s_{i,1}....s_{i,6}\}$ in (4) for the user to gauge the level of confidence in each response.
\begin{align}\label{eq1}
    Q_i=I_i \cup Q_{N_i}, \\
    Q_{N_i}=\cup_{k,k',k''}\{Q_{M_k}, Q_{G_k'}, Q_{T_k''}\},\\
    P_i=\cup \{\Psi_i,d_i,c_{opt,i}\},\\ 
    S=Score(R_i, \{Q_{M_k}, Q_{G_k'}, Q_{T_k''}\}, \{R_{M_k}, R_{G_k'}, R_{T_k''}\}).\\ \nonumber
\end{align}

\subsection{Data Chunk Ranking Module}\label{chunk}
The objective of this module is to prevent the LLM from being overwhelmed by massive amounts of data, and thereby running into token limitation errors. This module has three steps to correct for typing errors, then to filter the data chunks by keywords based on query $Q_i$ followed by selecting the data chunks that are most \textit{similar} to the query. These steps are motivated by the RAG methodology for LLMs in \cite{RAG} and described below.
\begin{enumerate}
\item In the first step, a Levenstein-distance based spelling checker is applied to the user-query to counteract manual typing errors. Next, a sophisticated keyword dictionary is used to filter data chunks that contain the financial metric, geo-location and time period in question (as $c_i$). This keyword dictionary contains details regarding financial metrics, geo-locations and time period inter-relationships and hierarchies.
\item In the second step, multiple searches are initiated to identify data chunks with the named entities ($Q_{N_i}$) while ensuring relational and hierarchical data integrity.
\item In the third step, the embedded version of the question ($Ada(Q_i)$) is combined with the filtered data chunks to identify the filtered data chunks with \textit{highest cosine similarity} \cite{survey}. Upto 20 top matching data chunks are returned to customize the prompt per user-query. This chunk limit of 20 is empirically set based on the token limitation of 4096 for the Azure GPT-3 model and can be increased as the LLM token limitations evolve. It is noteworthy that the business logic/rules in the first step are fine-tuned to ensure atleast one data chunk is matched to each query. In the absence of \textit{enough} data chunks, the LLMs tend to hallucinate and respond from their pre-trained data as shown in instance 3 of Fig. \ref{hallu}.
\end{enumerate}

\subsection{Custom Prompt Generation}
This module converts a simple user-query to a customized query with all relevant definitions and data to answer the question with minimum hallucinations. For our Finance-chatbot, the individual prompt templates corresponding to each question intention are composed of the following components: an introduction, a preamble, matching context, instructions and a few example questions and their answers. The details of each prompt component is explained as follows.
\subsubsection{Prompt Introduction}
 This section sets the persona for the LLM responses and sets up basic instructions for the LLM to understand key-definitions and the steps the LLM must follow to answer the question. This section explains the information flow in the prompt to the LLM and needs significant modifications while switching between LLM types.
\subsubsection{Prompt Preamble}
This section guides the LLM through domain-specific terminologies that are unique to the use-case. While there may be several custom definitions ($D$), only the definitions relevant to the named entities in the user-query are used each time ($d_i$). For niche domains such as Finance, there are often a wide variety of acronyms such as `PPP' that may stand for `Profit Per Period'. This definition is critical to extracting the right context from the data chunks. Thus, this section introduces all necessary definitions and data inter-relationships and hierarchies that are necessary to extract the necessary context from data chunks.

As an example, we generate a customized prompt for a user-query: \textit{`Where in Europe is the highest GDP growth in FY23?'} The process begins by gathering the most related data chunks based on Section \ref{chunk}. Next, the intention and named entities are identified as `GDP' and `Europe' from the question. The preamble includes relational details that mentions countries like `Germany,' `France,' and `UK' belong to `Europe' and hence data chunks from these countries are \textit{relevant}. Thus, the country names/keywords are intelligently mapped to the original named entities in the user-query through the preamble corpus. While the preamble component structurally resides next to the introduction section in the prompt, it is the last component to be executed by the custom prompt generation pipeline. This is due to its unique requirement of having both the user-query and contextual data in place prior to making the inter-connections dictated by the business logic specific to the use-case. Thus, the preamble section maximizes the prompting capabilities in domains that are heavy on custom definitions and jargon.
\subsubsection{Prompt Context}
This component combines all the matching data chunks and creates an overall context for the LLM to access while building a response. It is noteworthy that providing \textit{more} data chunks to the LLM is less detrimental than providing \textit{less} data chunks, since the LLMs typically have the capability to discern between sentences with specific named entities. For instance if a user-query is: ``Which countries in Asia have below average GDP growth in FY23?", then several data chunks with Asian country names and their GDP relative to the average Asian GDP are returned and the LLM selects the sentences that have lesser than average Asian GDP and paraphrases the response. In this scenario, passing few data chunks can lead to hallucinations due to LLM training biases. 
\subsubsection{Prompt Instructions}
This component conveys step-wise directions on how the LLM should formulate its response, while tailoring to the unique question intention category. These steps are based on the open-source Langchian framework in \cite{langchain}. For instance, fetching the answer to a simple user-query such as: ``What is the growth in USA in Q3 FY23?" would require fewer steps than the user-query: ``Why has the growth in USA slowed down last quarter?" or the query ``How can the growth in USA be improved in Q1 FY24?". For the simple ``What/where" queries, the goal is to find the most relevant sentences and to paraphrase them, whereas for the more complex ``Why/How?" questions, instructions dictated by the business logic and data relationships need to be laid out for the LLM to extract the most relevant context and to summarize and paraphrase them. It is noteworthy that each intent-specific prompt templates contain sequential, algorithmic instructions along with instructions to handle limiting conditions (when too many or too few data chunks are returned) to return coherent and contextually relevant responses.
\subsubsection{Prompt Example and Question}
This final component represents the precise format of the desired response. For instance, GPT models (GPT3/3.5) tend to generate responses in a paragraph format. Depending on the user interface and the volume of information comprehension, a paragraph format may not always be the desired response. Thus, for Finance-chatbots it is a good practise to specify a bullet-point format for responses. Additionally, the sample question and its response can be used as a guidance mechanism to include relevant sources and links in the response as well. One major caution while generating this segment is that a sample example need not be a precise example and its directed answer, but it can be a sequence of strategic questions and answers that aid interpretation of abstract user-queries while answering a board spectrum of questions in a prescribed format.

The overall process of customized prompt generation ($P_i$) per user-query $Q_i$ can be summarized in Algorithm \ref{algo}.
\begin{algorithm}[ht!]
\SetAlgoLined
\KwOut{Customized Prompt $P_i$}
 \KwIn {Query $Q_i=I_i \cup Q_{N_i}$}
 Initialization\;
     $I_{i} \leftarrow$ classify-intent($Q_i$),  $I_i \in [0,1,2,...l]$\\
     $\Psi_i \leftarrow$ Prompt-template($I_i$)\\
     $E_{Q_i} \leftarrow$ Ada-embedding($Q_i$)\\
     size$(c_{opt,i})$=0\\
\While{size$(c_{opt,i})<20$}{
$c_i \leftarrow$ Filter-data-chunks($C,Q_{N_i}$)\\
$c_{opt,i} \leftarrow c_{opt,i} \cup \arg\max(E_{Q_i}$, Ada-embedding($c_i$))\\
}
$d_i \leftarrow$ Filter($D, Q_{N_i}$)\\
$P_i =\cup \{\Psi_i, d_i, c_{opt,i}\}$
    \caption{Custom Prompt Generation}\label{algo}
     \vspace{-0.1cm}
\end{algorithm}

\subsection{Scoring Engine}
All previous modules described above are aimed at minimizing the LLM hallucinations prior to receiving a LLM response. However, based on the abstractness of the user-query and limitations in data chunks, hallucinations may still occur and they must be detected and reported to the users along with the LLM responses. To measure the quality of each LLM response, we propose a light-weight novel scoring module using `nltk-libraries' wherein the content of each response is scored using 6 binary metrics described below. 
\begin{enumerate}
    \item Question-answer continuity metric ($s_{i,1}$): This binary metric analyzes the named entities in both the question ($Q_{N_i}$) and response ($R_{N_i}$) texts. Based on this metric, we can infer if the response addresses the user-query or if the LLM has hallucinated and gone off track. For example, if the user-query is about ``quarterly costs figures”, and the LLM responds with details regarding ``annual figures”, then this metric is 0 as shown in \eqref{s1}.
    \item Numeric hallucination metric ($s_{i,2}$): When the model returns a response, it is essential to validate that the numbers in the response exist in the context section of the customized prompt. Otherwise, it is highly likely that the facts and figures are fabricated hallucinations. For example, if the LLM response includes: ``The metric ZZ has increased by X.X\%'', then the value X.X\% must be included in the prompt context. This binary metric is set to 1 if all the numbers in the response are a subset of the numbers in the prompt context as shown in (6).
    \item Uniqueness from prompt metric ($s_{i,3}$): This binary metric identifies if a consecutive sequence of $\delta$ words is the same within the response and the prompt. This metric evaluates the paraphrasing capability of the LLM to answer abstract and complex user-queries. This metric is set to 1 if there are no sequence of 10-12 words in a row in the response matching with the customized prompt as shown in \eqref{s3}.  \item Numeric sensibility metric ($s_{i,4}$): This binary metric checks for grammatical flaws in sentences involving numerical analysis. For instance, if the $\alpha$-th response sentence ($R_{i,\alpha}$) contains the word ``increase/increasing'', this sentence should not be accompanied by a negative number or percentage. Similar number and textual trends are analyzed for each response sentence. This metric is set to 1 only if all response sentences are grammatically sensible as in (8).
    \item Context warning metric ($s_{i,5}$): This warning metric detects if the context is easily accessible or not. In situations where multiple data chunks of different sizes and varying financial metrics are combined to generate a composite context, this metric is set to 1. This metric is useful in situations where users don't approve of a LLM response, wherein this metric can indicate erroneous data chunks being pulled into the customized prompt. 
    \item Contextual continuity metric ($s_{i,6}$): This metric analyzes each response sentence ($R_{i,\alpha}$) and verifies if the context (or named entities) match those in the data chunks of the prompt ($c_{opt,i}(N_{i,\alpha})$). For instance, if the LLM response contains, ``profits have increased by Y.Y\% in market$_1$'', but the prompt data chunks contain the sentence ``In quarter TT, the profit has increased by Y.Y\% in market$_2$", then this metric is set to 0 as shown in (9).
\end{enumerate}

\begin{align}\label{s1}
s_{i,1}=\left\{
        \begin{array}{ll}
        		1  & \mbox{if } Q_{N_i}==R_{N_i} \\
        		0 & otherwise \\
        \end{array}
\right\},\\ \nonumber
s_{i,2}=\left\{
        \begin{array}{ll}
		1  & \mbox{if } numbers(R_i) \subset numbers(c_{opt,i}) \\
		0 & otherwise
	\end{array}
 \right\}, \\
 \end{align}
 
\begin{align}\label{s3}
s_{i,3}=\left\{
        \begin{array}{ll}
        		1  & seq_{\delta}(P_i) \cap seq_{\delta}(R_i) ==\phi\\
        		0 & otherwise \\
        \end{array}
\right\},\\ 
s_{i,4}=\left\{
        \begin{array}{ll}
        		1  & ``positive" \in R_{i,\alpha}, numbers(R_{i,\alpha})>0\\
        		1 & ``negative" \in R_{i,\alpha}, numbers(R_{i,\alpha})<0\\
                    0& otherwise
        \end{array}
\right\},\\ \nonumber
\end{align}
\begin{align}
s_{i,6}=\left\{
        \begin{array}{ll}
        		1  & R_{N_{i,\alpha}}==c_{opt,i}({N_{i,\alpha}})\\
        		  0& otherwise
        \end{array}
\right\}.\\ \nonumber
\end{align}
Finally, based on the response scores, a confidence is associated to each LLM response and returned to the user as \{High, Medium, Low\} as shown in \eqref{confi}. It is noteworthy that the response confidence scores assist the user by telling them that the LLM either understood or completely failed to comprehend the user-query. A low confidence response is typically associated with open ended or abstract user-queries, inadequate data chunks, incorrect question intention classification and hallucinations. The confidence score tells the user to assert caution while making key decisions using medium to low confidence responses. The confidence score further helps ascertain which user-queries need to be further refined for reliability.
\begin{align} \label{confi}
Confidence(R_i)=\left\{
        \begin{array}{ll}
        		High  & \sum_j s_{i,j}>=5\\
        		  Medium& 4>=\sum_j s_{i,j}>=3\\
                    Low & otherwise
        \end{array}
\right\}\\ \nonumber
\end{align}

\subsection{System Architecture for Finance-chatbot}
\begin{figure}[ht]
  \centering
  \includegraphics[width=\linewidth]{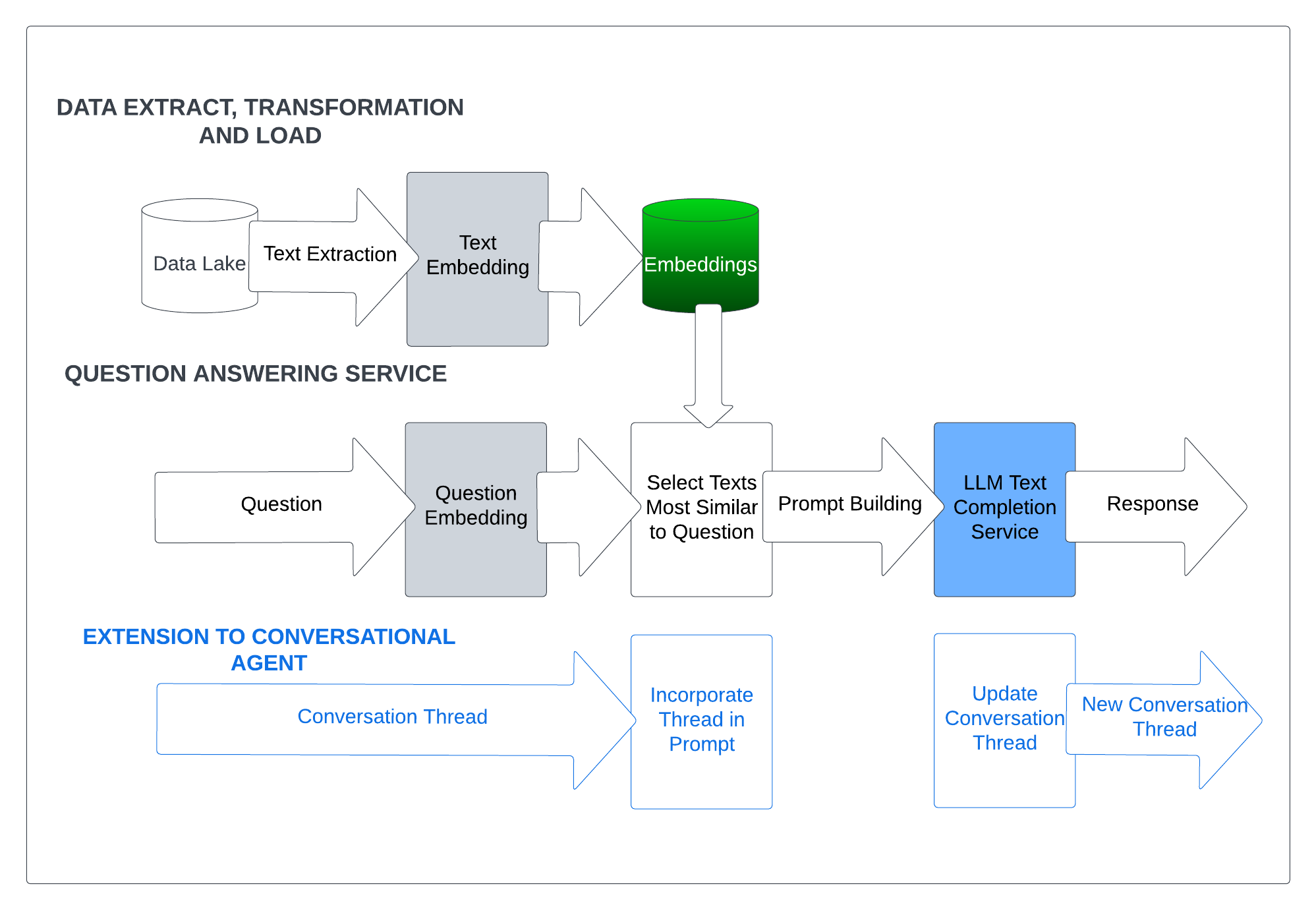}
  \caption{System diagram of the scalable Data-to-answers setup with conversational threads.}\label{scale}
\end{figure}

The major differences between an LLM prototype and a product include the following: scalability in infrastructure with growing data chunks, robustness to variations in user-queries, data freshness, security concerns and constant enhancements based on human feedback. Thus, the following considerations are required to design a Finance-chatbot at an Enterprise-level.
\begin{itemize}
    \item Data tables must be refreshed based on Enterprise-level guidelines to ensure reliable LLM responses.
    \item Security and access constraints at user level needs to be implemented at data chunk storage level.
    \item Fewer LLM calls per user-query  may prevent hallucinations while ensuring fast response times. However, multiple LLM calls per user-query to classify user-query intentions, to extract named entities or match the best data chunks may enhance response performances for previously unseen and abstract user-queries.  
    \item Custom prompt generation with best matching data chunks is one way of implementing RAG \cite{RAG} to extract customized responses by grounding. However, additional user-history based information may further enhance the question answering experience at a personalized level.
    \item Building chatbots for finance-based decision makers involves reliable and repeatable response generation from the LLM along with the response scores that guide the users/decision makers. 
    \item Feedback collection is an integral part of the product pipeline to ensure that ``low'' confidence responses are constantly recorded and improved based on business logic.
\end{itemize}
To ensure a scalable and usable product, the system architecture in Fig. \ref {sys} can further be expanded to work as a conversational agent as shown in Fig. \ref{scale}. This can be achieved by including the conversation thread in the input of the live process service that initiates when each user-query is entered. Passing the conversational continuation information to the custom prompt enables answering follow-up questions using the previously selected data chunks. This creates a conversational thread that is constantly updated with each LLM prompt and the returned responses. This design allows for an external calling application to orchestrate a conversation with follow up queries.

Another design consideration for a scalable personalized chatbot is the data Extract, Transformation and Load (ETL) functions \cite{etl} to generate data chunks following each data refresh. This offline process can be further modified to allow for almost-real time data chunk generation, by incorporating an asynchronous publish/subscribe system where source data changes are published, and the data chunk creation system subscribes to these changes to generate new updated data chunks \cite{architecture}. Some existing alternatives to the data chunk ranking mechanism are ElasticSearch, Apache Solr or Lucene \cite{elastic}.

One major consideration for scaling a Finance-chatbot as a service is horizontal scaling to support concurrent users by running parallel service instances. Here, we assume that the user-query intention classification and custom prompt generation services also scale in parallel. As data grows, the main bottleneck is the time taken for keyword matching while ranking the best matching data chunks. From a platform perspective, if the keyword matching process is designed around an SQL database, then indexing must be improved to maintain fast response times ($<3$ seconds  per question). However, for extremely large amounts of data (like few hundred thousand to a million data chunks) database (horizontal) sharding needs to be considered \cite{sharding} wherein, access to data chunks is rerouted based on a metric or dimension. ETL processing times can be a secondary bottleneck in such situations with growing data chunks. By updating from a batch ETL to an almost real-time data chunk generation process as described above the overall query processing times can further be stabilized.

Finally, we can compare our proposed Finance-chatbot architecture in Fig \ref{scale} with the more common RAG architecture in \cite{RAG} to answer questions based on domain-related unstructured data as well. For instance, if we had to extend the Finance-chatbot to learn from fiscal meeting call summaries and minutes of meetings, then the following architectural scaling are necessary. First, the major difference in such a scenario is that instead of matching questions to data chunks using keywords followed by filtering and ranking the data chunks for a customizable prompt generation, we rely on semantic similarity (of the Ada002 embeddings \cite{ada}) between questions and text fragments to filter and rank. The same considerations for scaling as discussed above hold. However, sharding a vector database may require a separation between topics for such unstructured datasets, rather than using the structured tabular dimensions. Thus, the infrastructure scaling considerations discussed here can be applied to structured and unstructured datasets.

\section{Experiments and Results}
In this work, we perform 4 major experiments to ensure reliability, repeatability and trustworthy responses from our Data-to-answers Finance-chatbot. First, we assess the best practices for custom prompt generation required to minimize hallucinations while being LLM agnostic. Second, we assess the GPT-3/GPT3.5 LLM response quality scores across a batch of over 350,000 curated user-queries. Third, we assess the importance of user-query intention classification using a separate LLM prompt as opposed to a standard pre-trained classifier. Finally, we assess the response performance variations across a variety of LLMs for benchmarking purposes.
\subsection{Prompt Engineering for minimized Hallucinations}
The best practices for advanced prompt engineering to control for corner conditions and to ensure reliable and repeatable responses are discussed below:
\begin{enumerate}
    \item Hallucination guardrails: Using specific text in the prompt introduction and preamble sections that instructs the LLM to refrain from generating responses when uncertain, or when there is a lack of contextual data. 
    \item Modular custom prompts: Creating modular prompts enables adaptability to new user-query intentions, prompt template re-usability and scalability to abstract user-queries.
    \item Customization as needed: Tailoring prompts to suit the particular requirements of a task or query is imperative for achieving coherent and reliable responses. Customization allows for accommodating business logic and data inter-relationships as needed for specific query intentions such as to explain ``Why?'' situations. 
    \item Detailed instructions: Any custom prompt must be as detailed with easy to follow steps and guidelines. Step by step instructions enable debugging and troubleshooting to minimize hallucinations. However, instructions should not be too long sentences in which case LLM token limitations may get triggered.
    \item Advanced prompting: Prompt instructions should enable handling for limiting conditions, for instance when few data chunks or a short context is available the response must only answer from the context or return ``I cannot answer the question.'' Also, in scenarios when the user-queries are abstract with multiple interpretations, clear guidelines must be provided to respond with cautionary text. For instance if the user-query asks about the ``GDP in north eastern states in USA'' while the context contains aggregated USA data only, the LLM must be prompted to respond with text such as ``Apologies, but I can only answer with regards to the aggregated GDP data form USA.'' Disambiguation enables higher reliability on the response even if it is not exactly what the user wanted.
    \item Use of delimiters: Incorporating delimiters within the prompt, such as '---text---' or $'<$text$>'$, aids the LLM to identify the location of specific information such as definitions and instructions, thereby improving response accuracy.
    \item Avoid contradictory instructions: Ensuring that prompt instructions are consistent while avoiding contradictions is necessary in limiting conditions to avoid hallucinations.
    \item Prompt evolution: The process of custom prompting is iterative with a focus on performance over perfection. As LLMs get fine-tuned based on the usage, they may support larger context in the near future. However, there will always be a need to fine-tune and evolve the LLMs to enable complex decision making tasks such as suggesting prescriptive trends and scenario planning capabilities. Evolving the prompts based on human feedback is a necessary component to enable enhanced user experiences.
\end{enumerate}

\subsection{LLM Response Quality Scoring}
\begin{figure}[ht]
  \centering
  \includegraphics[width=\linewidth, height=2.5in]{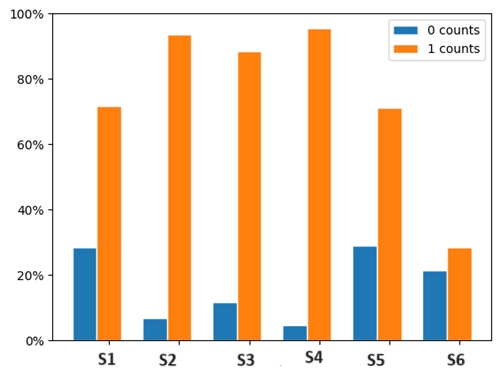}
  \caption{The average LLM response scores across a range of user-queries.}\label{bar}
\end{figure}

In Fig. \ref{bar}, we observe the average LLM response scoring metrics $\{s_{1},...s{6}\}$ on a list of \{$i=1:356,000$\} curated set of user-queries that range across all intention categories.
Since all the response scoring metrics are binary in nature, a higher frequency of 1 values reflect good response statistics per user-query and 0 values reflect possible hallucinations. Here, we observe that the average scores for metrics $\{s_{2}, s_{3}, s_{4}\}$ are the highest. This implies that the Azure GPT-3 (davinci-003 legacy) LLM hallucinates fake numbers less than 10\% of the time, as detected by $s_{i,2}$ in (6). Also metrics $\{s_1, s_6\}$ focus on named entity searches and here we observe over 20\% rate of ambiguities in the response contexts. Also, we observe that the warning metric $s_{5}$ has the highest false rate of over 30\%, indicating that the data chunks selected in the context contain a variety of finance metrics that are different from the finance metric in the user-query. While a high false rate for metric $s_{5}$ does not necessarily imply hallucinations, this metric serves as a diagnostic tool to improve question responses based on human feedback since it is indicative of issues with the data chunk filtering and selection modules. These average response scores for an offline set of user-queries enable quality assurance for versioned deployments of the Finance-chatbot.  

\begin{figure}[ht]
  \centering
  \includegraphics[width=\linewidth, height=2.5in]{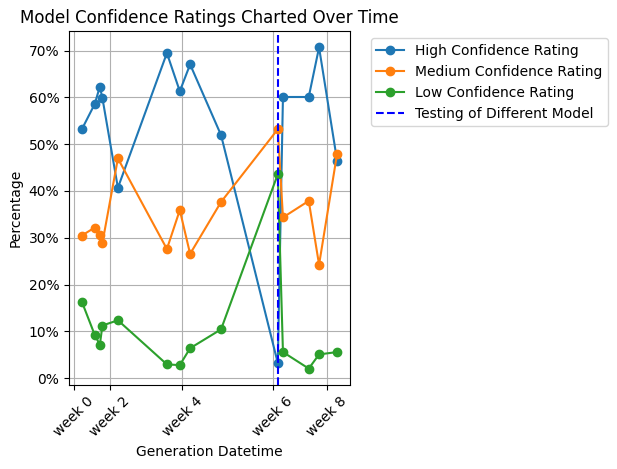}
  \caption{Variations in average response confidences per weekly deployment assessed on a curated set of 356,000 user-queries. The week 6 deployment is indicative of switching LLMs from GPT-3 (davinci-003 legacy) to GPT 3.5.}\label{conf}
\end{figure}

Finally, we assess the overall confidence scores generated for the same batch of 356,000 questions by assessing the combined impact of the quality scoring metrics. The response scores per user-query is categorized as \{Low/Medium/High\} based on (10) and variations in the average response confidences across weekly deployments over a period of 8 weeks is shown in Fig. \ref{conf}. The response confidence scores are indicative if the LLM understood the user-query and if it had the required data to furnish a reliable answer. While a ``low'' confidence may not necessarily imply a hallucination has occured, it offers the user an additional reliability criteria to do their due diligence and arrive at the answer from other sources as well. The response confidence scores also reflect an evolution in the LLMs and advanced prompting capabilities over time. Additionally, the confidence scores aid prioritization for prompt enhancements to enable iterative evolution in overall LLM response quality, reliability and trustworthiness.
In Fig. \ref{conf}, we observe a consistent improvement in overall confidence scores by iterative advanced prompting. The major change in response confidence ratings at week 6 time-frame indicates a change in LLM from GPT3- GPT 3.5. This observation indicates that while most Langchain frameworks \cite{langchain} are LLM agnostic, the prompts need to be re-engineered to a certain extent for every LLM change. Thus, the overall response scoring and confidence measurement modules are significantly important to plan and monitor such back-end LLM changes that may otherwise impact user-experience significantly. 

\subsection{End-to-end Data to Answers with Multiple Prompts}
The first and major component in our Finance-chatbot that controls for hallucinations is the intention classification process for each user-query. While this process can be easily automated by a supervised learning models such as decision trees, one vs. all linear classifiers, SVM etc, classical classifiers are sensitive to keywords used in the query and noisy data entry caused by typing errors or abstract user-queries can cause a pre-trained classifier to mis-classify, leading to the incorrect prompt template being used for customized prompt generation thereby leading to a low/medium quality response that the user may dislike. However, LLMs have been used successfully for simple sentiment classification tasks and they are robust to typos as well \cite{finbert}. In this experiment we evaluate the one vs. all classification performance of a linear kernel SVM model that is pre-trained on 900,000 user-queries that are categorized into 9 classes. The SVM classifier has the best test scores on a 70/30 split for the 900,000 user-queries when compared to decision trees, XGBoost and linear classifiers. We compare the SVM classification performance (in terms of average precision (Pr) and recall (Re)) with that of an LLM prompt that passes instructions to read the user-query and to classify it into one of the 9 intention categories. For test data, we curate a set of 50 open ended user-queries that are abstract in nature. The intent classification performances for each category of user-queries using the SVM classifier vs. a custom LLM prompt are shown in Table \ref{e2e}.
\begin{table*}[ht] 
\caption{Classification performances of user-query classification using SVM  vs. a custom LLM prompt.}
\scalebox{0.9}
    {
\begin{tabular}{|c|c|c|c|c|}
\hline
{\bf Intent Category}&{\bf Question Types}&{\bf Example user-queries}&{\bf SVM Pr/Re}&{\bf Prompt Pr/Re}\\
\hline
0&Basic Information, Definitions&\textit{What is the growth in USA in FY 23?}&0.714/0.625&1/0.625\\\hline
1& Ranking (Highest/Lowest)&\textit{Which regions in USA have the highest revenue?}&0.446/0.7&0.45/1\\ \hline
2& Direction (Increasing/Decreasing)&\textit{Is the revenue in Canada increasing?}&1/0.33&1/0.83\\ \hline
3&General insights and summaries&\textit{Summarize the key economic insights in USA?}&0.8/1&1/0.5\\ \hline
4&Problem-Solving&\textit{How can the agricultural revenue be improved in USA?}&0.75/1&1/1\\ \hline
5& Diagnostics& \textit{What are the top drivers for revenue deficits in the Midwest in FY23?}&1/1&1/1 \\\hline
6& Performance&\textit{How is the revenue trend in the south for industrial products in FY23?}&0.5/1&1/0.67 \\\hline
7& Outliers&\textit{What are the outliers/exceptions for financial stocks in NYSE in FY23?}&1/0.75&1/1 \\ \hline
8& Impact&\textit{How does the revenue deficits for northeast impact the Midwest trends?}&0/0&1/0.4\\ \hline
\end{tabular}\label{e2e}
}
\end{table*}
Here, we observe that the macro-level Pr/Re for LLM-based intent classification is 0.94/0.78, which is significantly higher than the macro level performance statics for the SVM classifier Pre/Re of 0.69/0.71. This observation aligns with the prior work in \cite{Sun23}, wherein GPT-3 is shown to have better text-based classification performances when compared to standard classifiers. Thus, using multiple LLM prompts to first categorize intention of user-queries followed by building a custom prompt with relevant context, definitions, instructions and examples can significantly reduce hallucinations. However, it is noteworthy that multiple LLM prompts per user-query may increase the response times. 
 \subsection{Benchmarking for LLM Performances}
Designing LLM-based products requires four major considerations that are different from the prototyping stage. These considerations are response reliability, response time, infrastructure security and usage costs. 
Although our system design for the Finance-chatbot has been LLM agnostic so far, we need to constantly weigh the potential risks and advantages offered by a variety of LLMs to maintain the reliability and accuracy of responses. In this experiment, we present a comparative assessment of the pros vs. the cons of LLM changes and our LLM selection criteria.
\begin{table*}[ht] 
\caption{Analysis of response variations based on LLM changes.}
\scalebox{0.82}
    {
\begin{tabular}{|c|c|c|c|c|c|c|c|c|}
\hline
{\bf LLM name}&{\bf Prompt token Cost (\$)}&{\bf Token Limit}&{\bf $\bar{s_6}$}&{\bf $\bar{s_2}$}&{\bf$\bar{s_3}$}&{\bf$\bar{s_4}$}&{\bf Pros}&{\bf Cons}\\\hline
GPT3 (legacy)&0.005&4096&0.83&0.93&0.86&0.93&Creative Responses&Higher cost, hallucinations\\ \hline
GPT3.5&0.0008&4096&0.89&0.93&0.69&0.93&Lower cost, Conversation memory&Higher hallucinations that GPT3\\ \hline
Text-Bison&0.0005&8192&0.92&0.91&0.90&0.95&Lower cost, lower hallucinations&No additional context for follow ups.\\ \hline
Chat-Bison&0.0005&4096&0.85&0.72&0.88&0.95&Lower cost, Conversation memory&Hallucinations for big contexts only\\ \hline
TextBision 32K&0.0005&32000&0.92&0.91&0.91&0.95&Lower cost, high token limit, lower hallucinations&Higher overall cost per query\\ \hline
\end{tabular}\label{LLMs}
\vspace{-0.1cm}
}
\end{table*}

For this experiment, we apply a carefully curated set of 300 user-queries corresponding to the 9 intention categories. We select straightforward and open ended questions for our assessment. The assessment regarding the LLM cost incurred per token and the average response scores for the 300 user-queries along with the overall pro and con of each LLM is shown in Table \ref{LLMs}. Here, we observe that the Bison 32K model has the highest response scores ($>90\%$) while the Chat-Bison and GPT 3.5 have the least response scores. This observation is intuitive since Chat-Bison and GPT3.5 have been trained specifically to maintain conversations rather than return summaries for large contextual data. Thus the optimal LLM selection plan would be between GPT-3 annd Text-Bison for such Data-to-answers use-cases. 

\section{Conclusions and Discussion}
LLM-powered chatbot solutions for niche domains such as Finance are faced with major challenges such as reliability and scalability. In this work, we present a novel LLM agnostic framework that aims to generate responses to a variety of user-queries from pure data tables while controlling for hallucinations in the responses. Our journey from prototyping to product scaling has provided several leanings to maximize custom prompt generation capabilities while ensuring minimized hallucinations for financial decision makers. Our novel framework controls for hallucinations or fake LLM responses at various levels and it also scores each response to provide an additional level of confidence to the users regarding the accuracy of the response. The proposed system setup is significantly dissimilar from generic question answering chat-bots since small error rates and hallucinations can have detrimental impacts for financial decision makers. The proposed response scoring module ensures that the users comprehend a level of acceptance for the LLM-generated responses, thereby detecting the 10\% instances when the response scores are ``low'' that may be indicative of potential hallucinations. The proposed system has been evaluated for versioned deployments and for variations in LLM providers as well.

The four major conclusions from this work are as follows. First, we observe that advanced prompt engineering and modular infrastructure design plays a crucial role in ensuring reliable and accurate responses. Second, the proposed novel response scoring module enables response quality monitoring between data-refreshed deployments. Third, we investigate the use of multiple LLM prompts per user-query to ensure higher reliability and lower hallucinations in responses. We observe that using LLMs have robust user-query intention classification performances when compared to standard SVM classifiers for relatively abstract user-queries. Also LLM-based intention classification is more robust to noisy data and typing errors, thereby making multi-prompt question-answering systems more robust than single prompt frameworks. Fourth, we present our LLM benchmarks and criteria for selection of the LLMs for domain-specific use cases based on cost-per-query, reliability and capability to scale.

The novel Data-to-answers framework presented in this work demonstrates the system and infrastructure level considerations required to take a Finance-chatbot from prototyping to scaled stages. The proposed framework has the following advantages over standard Langchain frameworks.
\begin{enumerate}
    \item Versioned deployments to ensure evolution in advanced prompting.
    \item Ease of use to build reliability and trustworthiness for the users to enable financial decision making.
    \item Infrastructure scalability to support increasing data-driven capabilities by incorporating data predictions and trend-related information into data chunks.
    \item LLM agnostic framework that can be monitored for response performances to ensure evolution in response quality with human feedback.
\end{enumerate}
Future works will be directed towards further enhancing the hallucination minimization frameworks for unstructured financial data and documents combined.
\section*{Acknowledgment}
The authors would like Oscar Rodrigiuz, Kiranjit Kaur, Juan J. Ricas Corona, Errol Paclibar and Anandam Dhandugari for their constant experimentation support. Special thanks to Priya Raman for constant guidance and support.

\bibliographystyle{IEEEtran}
\bibliography{sample}

\end{document}